\documentclass[10pt,twocolumn,letterpaper]{article}

\usepackage{cvpr}
\usepackage[titletoc,title]{appendix}
\usepackage{times}
\usepackage{subcaption}
\usepackage{epsfig}
\usepackage{graphicx}
\usepackage{amsmath}
\usepackage{amssymb}

\graphicspath{{./graph/}{./}}

\cvprfinalcopy 

\def\cvprPaperID{****} 

\begin{document}

\title{The Monkeytyping Solution to the YouTube-8M Video Understanding Challenge}

\author{He-Da Wang \\
{\tt\small whd.thu@gmail.com}\\
\and
Teng Zhang\\
{\tt\small zhangteng1887@gmail.com}\\
\and
Ji Wu\\
{\tt\small wuji\_ee@mail.tsinghua.edu.cn}\\
Multimedia Signal and Intelligent 
Information Processing Laboratory\\
Department of Electronic Engineering\\
Tsinghua University, Beijing, China\\
}

\maketitle

\begin{abstract}
   This article describes the final solution \footnote{visit https://github.com/wangheda/youtube-8m for the source code.} of team monkeytyping, who finished in second place in the YouTube-8M video understanding challenge. The dataset used in this challenge is a large-scale benchmark for multi-label video classification. We extend the work in \cite{Youtube-8M} and propose several improvements for frame sequence modeling. We propose a network structure called Chaining that can better capture the interactions between labels. Also, we report our approaches in dealing with multi-scale information and attention pooling. In addition, We find that using the output of model ensemble as a side target in training can boost single model performance. We report our experiments in bagging, boosting, cascade, and stacking, and propose a stacking algorithm called attention weighted stacking. Our final submission is an ensemble that consists of 74 sub models, all of which are listed in the appendix. 
\end{abstract}

\section{Introduction} \label{Section:Introduction}

Videos have been a very important type of content on Internet. Understanding video from its audio-visual content is key to various applications such as recommendation, searching, and question answering. The research on video analysis is also an important step for the computer to understand the real world. Datasets such as Sports-1M\cite{largescale2014karpathy} and ActivityNet\cite{activitynet2015caba} encourage the research on video classification of sports and human activities. YouTube-8M\cite{Youtube-8M} is a large-scale video dataset that consists of about 7.0 million YouTube videos that was annotated with a vocabulary of 4716 tags from 24 diverse categories. The average number of tags per video is 3.4.

The YouTube-8M video understanding challenge is an arena for video classification researchers and practitioners. In the competition, the dataset is divided into three parts. The training set contains 4.9 million samples. The validate set contains 1.4 million samples. The test set contains 0.7 million samples. The ground truth labels annotated to the samples in the training set and the validate set are available to the participants. The test set is divided into two half, the open test set and the blind one. In the progress of the competition, participants can submit their predictions to the test set. The scoring server on Kaggle would evaluate them and return the scores on the open test set (the public leaderboard). After the competition, the winners are decided by the scores of their submissions on the blind test set (the private leaderboard).

In the competition, submissions are evaluated using Global Average Precision (GAP) at 20. The metrics is calculated as follows. For each video, the most confident 20 label predictions are selected along with the confidence values. The tuples of the form $\left\lbrace video, label, confidence \right\rbrace$ from all the videos are then put into a long list sorted by confidence values. This list of predictions are then evaluated with the Average Precision (Eq. \ref{Eq:AveragePrecision}), in which $p(i)$ is the precision and $r(i)$ is the recall given the first $i$ predictions.

\begin{equation}\label{Eq:AveragePrecision}
AP = \sum_{i=1}^{N} p(i) \Delta r(i) 
\end{equation}

\begin{figure*}[!htbp]
	\begin{center}
		\includegraphics[width=0.95\textwidth]{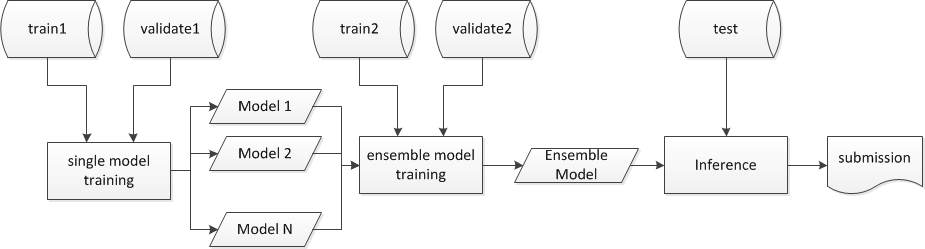}
	\end{center}
	\caption{The structure of our system.}
	\label{Fig:System}
\end{figure*}

We divide the dataset into five parts: train1, validate1, train2, validate2, and test. The division is done base on file name pattern (Table \ref{Table:DatasetDivision}). The set train1 is used for single model training, in which the set validate1 served as a hold-out test set for early-stopping. Ensemble models are trained on the set train2, which have no intersection with the training set of single models. The set validate2 is a hold-out test set for early-stopping in the training of ensemble models. The inference procedure uses the test set to generate submissions. The whole system structure is shown in Fig. \ref{Fig:System}. This data division strategy may not be optimal, since it limit the amount of data used in single model training to the 4.9 millions training set examples, which may considerably affect the performance of single models.

\begin{table}
	\begin{center}
		\begin{tabular}{|c|c|r|}
			\hline
			Part & File glob & \#samples \\
			\hline\hline
			train1 & train??.tfrecord & 4,906,660\\
			validate1 & validate[a]?.tfrecord & 21,918 \\
			train2 & validate[\textasciicircum a0-9]?.tfrecord & 1,270,412 \\
			validate2 & validate[0-9]?.tfrecord & 109,498 \\
			test & test??.tfrecord & 700,640 \\
			\hline
		\end{tabular}
	\end{center}
	\caption{Dataset division scheme in our system.}
	\label{Table:DatasetDivision}
\end{table}

During the training of all our models, We did not use any data augmentation techniques. Adam \cite{Adam2014Kinma} optimization algorithm is used throughout the training process. For models that use frame level feature $rgb,audio$, we use a learning rate of 0.001 and a batch size of 128. For models that use video level feature $mean\_rgb,mean\_audio$, the default learning rate and batch size is 0.01 and 1024. Note that these hyper parameters may not be optimal since we did not perform any parameter search. Every of our models can fit in the graphics memory of either an Nvidia GTX 1080 (8G) or an GTX 1080Ti (11G).

In the rest of this report, we summarize our contributions in our solution. We first introduce our improvements over the baseline Long-Short Term Memory (LSTM) model (Section \ref{Section:Baseline}). We then introduce a deep network structure that we use in many models to capture label correlations (Section \ref{Section:Chaining}). Then, we introduce our best performing single model that utilize multi\-scale information via Convolutional Neural Network (CNN) and LSTM (Section \ref{Section:MultiScale}). Also, we explore the use of attention pooling in video sequence classification (Section \ref{Section:Attention}).  Ensemble methods such as bagging, boosting (Section \ref{Section:BaggingBoosting}), cascade (Section \ref{Section:Cascade}), and stacking (Section \ref{Section:Stacking}) are also explored in this report. We also found that using the predictions of ensemble models as a soft target can greatly boost single model performance (Section \ref{Section:Distill}). In the end, we summarize our road-map and discuss the contribution of our solution(Section \ref{Section:Conclusion}).  

\section{Baseline models} \label{Section:Baseline}

In this section, we list the performances of some of the baseline models in the technical report \cite{Youtube-8M} and some alternative architectures evaluated in our system. The performances reported here are different from the ones reported in the technical report, since the dataset used in the report is a little larger (8.3 million examples) than the one provided in this competition. Also, we have limited the amount of data used for single model training to 4.9 million examples. The GAP scores we reported in this paper are evaluated with the "validate2" set if not otherwise specified.


We re-implemented four baseline approaches: Logistic Regression (LR), Mixture-of-Expert (MoE), Deep Bag-of-Frames (DBoF), and Long-Short Term Memory (LSTM). The GAP scores are listed in Table \ref{Table:BaselinePerformace}. We choose a mixture of 16 for the MoE model for the best performance. For the LSTM model, we choose the number of layers, the number of cells per layer, and the number of Moe mixtures to be 2, 1024 and 8 respectively.

\begin{table}[h] 
	\begin{center}
		\begin{tabular}{|l|l|c|}
			\hline
			Input Feature & Model & GAP \\
			\hline\hline
			Video Level, $\mu$ & Logistic Regression & 0.7577\\
			Video Level, $\mu$ & Mixture-of-Experts & 0.7957 \\
			\hline\hline
			Frame Level, $\{x^v_{1:F_v}\}$ & Deep Bag-of-Frames & 0.7845 \\
			Frame Level, $\{x^v_{1:F_v}\}$ & LSTM & \textbf{0.8131} \\
			\hline
		\end{tabular}
	\end{center}
	\caption{The performance of baseline approaches in our system.}
	\label{Table:BaselinePerformace}
\end{table}


In the conventional LSTM \cite{LSTM1997hochreiter}, the memory cells that share the same input gate and the same output gate form a memory block. Putting more than one cells into a block makes the training of LSTM more efficient. We create an architecture in which all the cells share a single input gate and a single forget gate and call it "LSTM-S". In this architecture, the cells use individual output gates, since we find that sharing the output gate can be harmful to the performance. 

Another network structure we create by adding an input accumulator to the architecture of LSTM model is what we call the "LSTM-A". In this structure, we add a new set of memory, input gate and forget gate to the cell to directly "remember" the input. Eq. \ref{Eq:LSTMA} gives the architecture of this model. In the equation, $x_t, c_t, h_t$ are the input, the memory, and the hidden state, $o_t,i_t,f_t$ are the output gate, the input gate, and the forget gate, $c'_t,d_t$ are the added memory and the hidden state for input accumulation, $i'_t, f'_t$ are the added input gate and forget gate, $\sigma$ is the sigmoid function, $g$ is the activation function which we choose to be $tanh$, and $n$ is the L2-normaliza\-tion function.

\begin{equation}\label{Eq:LSTMA}
	\begin{array}{cll}
		\left(
			\begin{array}{l}
				o_t\\
				m_t\\
				i_t\\
				f_t\\
				i'_t\\
				f'_t
			\end{array} 
		\right) & = & T_{N+2M, 6N} \left(
			\begin{array}{l}
				h_{t-1}\\
				x_t\\
				d_{t-1}
			\end{array} 
		\right)\\
		c_t & = & \sigma(f_t) \cdot c_{t-1} + \sigma(i_t) \cdot g(m_t)\\
		h_t & = & \sigma(o_t) \cdot g(c_{t})\\
		c'_t & = & \sigma(f'_t) \cdot c'_{t-1} + \sigma(i'_t) \cdot x_t\\
		d_t & = & n(c'_t)\\
	\end{array}
\end{equation}

The performance of the two modified LSTM models are listed in Table \ref{Table:ModifiedLSTM}. The number of cells per layer are 1024. The number of mixtures in MoE is set to 8. The number of layers $l$ is shown in the table. The two models performs better than the original model and contribute to the ensemble because their varieties in model structure.

\begin{table}[h] 
	\begin{center}
		\begin{tabular}{|l|c|}
			\hline
			Model & GAP \\
			\hline\hline
			vanilla LSTM, $l=1$ & 0.8091 \\
			vanilla LSTM, $l=2$ & 0.8131 \\
			\hline\hline
			LSTM-S, $l=1$ & 0.8123 \\
			LSTM-A, $l=1$ & 0.8131 \\
			\hline
		\end{tabular}
	\end{center}
	\caption{The performances of the single-block LSTM (LSTM-S) and the input-accumulator LSTM (LSTM-A), compared to the original LSTM model.}
	\label{Table:ModifiedLSTM}
\end{table}


In the dataset, an example has both visual feature and audio feature. The naive way of modeling two different features is to concatenate them into a single feature. We find this to be questionable for two reasons. First, in many videos such as music videos and family albums, the audio and the visual content are independent of each other. Also, people often can make sense of a video only by its visual content or audio content. Therefore, the visual or audio activities do not have to happen spontaneously to be meaningful. We adopt a parallel way that model visual and audio content with separate LSTM models. The final states of the two LSTM models are then concatenated and go through an MoE unit. The performance of the parallel model is shown in Table \ref{Table:ParallelLSTM}. The number of layers is 2. The number of cells $c$ and the number of mixtures $m$ in MoE model are shown in the table. 

\begin{table}[h] 
	\begin{center}
		\begin{tabular}{|l|c|}
			\hline
			Model & GAP \\
			\hline\hline
			vanilla LSTM, $c=1024,m=8$ & 0.8131 \\
			\hline\hline
			parallel LSTM, $c_v=1024,c_a=128,m=8$ & 0.8161 \\
			\hline
		\end{tabular}
	\end{center}
	\caption{Parallel modeling of visual and audio content is better than simple feature concatenation.}
	\label{Table:ParallelLSTM}
\end{table}

There is a potential weakness in modeling visual and audio features independently. It might be preferable to allow the visual network and the audio network to interact at certain points, since a visual / audio event may be related to a former audio / visual event in the same video. We leave this issue to future works.

\section{Label Correlation} \label{Section:Chaining}


In multi-label classification settings, an example may be annotated many labels. Some labels tend to appear in the same example at the same time, some tend not to. Such information can be used to improve the performance of multi-label classification models. 


Classifier chain \cite{CC2009Read} create a chain of classifiers. One classifier on the chain predict one label by using not only the input feature but also the predictions of other labels from the previous models on the chain. The training of one chain of classifier involves training $L$ models in which $L$ is the size of the vocabulary of all labels. However, in real world problems where the direction of dependency is unknown, an ensemble of classifier chains is usually used, which makes the computation complexity even higher and intractable if the vocabulary of labels is large. A neural network structure that mimic the classifier chain \cite{CCDeep2014Read} is proposed to address the multi-label problem. However, such network structure has a depth of $L$ that makes it hard to optimize if the vocabulary of labels is large. 

\begin{figure*}[!htbp]
	\begin{center}
		\begin{subfigure}{0.3\textwidth}
			\includegraphics[width=\textwidth]{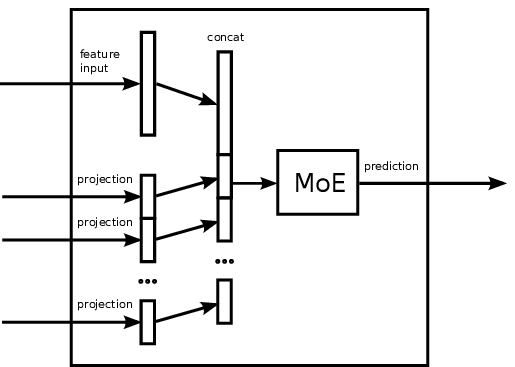}
			\caption{A unit in Chaining accept one feature and several model predictions as the input. The predictions are projected to lower dimension for efficiency.} \label{Fig:Chaining-Unit}
		\end{subfigure} 
		\hspace{0.03\textwidth}
		\begin{subfigure}{0.6\textwidth}
			\includegraphics[width=\textwidth]{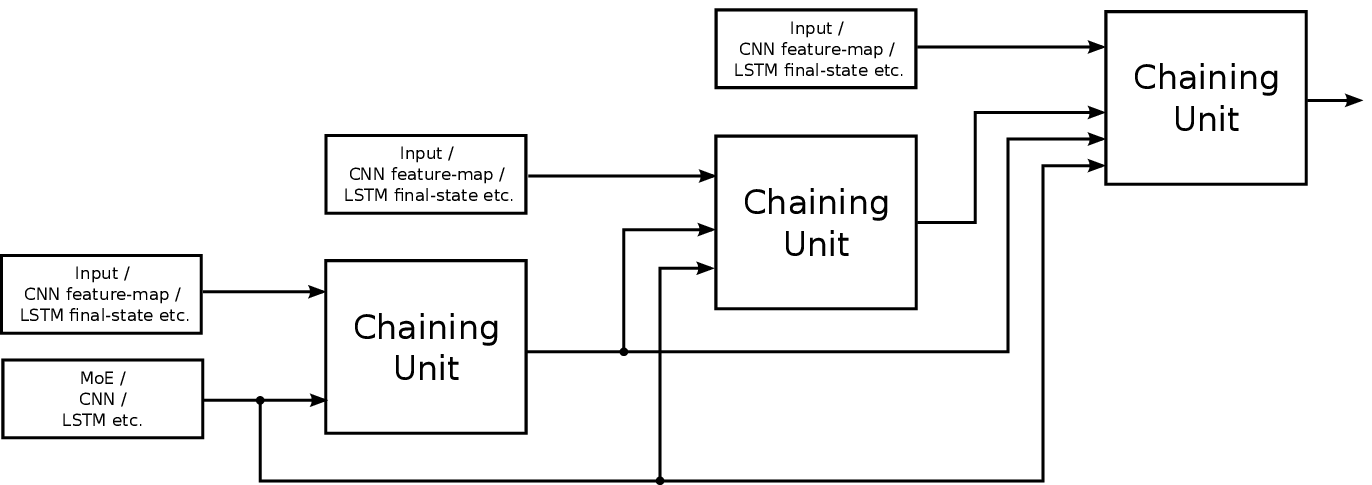}
			\caption{The whole architecture of Chaining consists of several stages. A Chaining unit in one stage accepts a feature vector, either from input directly or from a representation of LSTM or CNN, and all the predictions from earlier stages. } \label{Fig:Chaining-Graph}
		\end{subfigure}
	\end{center}
	\caption{Chaining: a deep learning architecture for multi-label classification.}
	\label{Fig:Chaining}
\end{figure*}

We propose a novel end-to-end deep learning structure that we called "Chaining" to better utilize the correlation between labels. In a Chaining model (Figure \ref{Fig:Chaining}), several representations are joined by a chain of MoE model. The predictions are projected to features of lower dimension and used in the following stages. The representations can be generated by homogeneous models or heterogeneous ones. We constantly apply auxiliary cross-entropy loss on the intermediate predictions to accelerate the training progress. The final loss function is a weighted average over the loss on the final prediction and the auxiliary losses. We typically allocate only $10 \% \sim 20 \%$ of the weights to the auxiliary losses, since the loss function at the final stage is the most important.


We performed experiments using three basic models, Mixture-of-Expert, LSTM and CNN. The CNN model we use in our system is the same as the benchmark method of sentence classification \cite{CNN2014Kim}, where the length of the filter is fixed to the size of feature vector per frame and the resulting feature map goes through a max-over-time pooling. The final state of LSTM and the max-pooled feature map are used as feature representation in Chaining models and go through an MoE model for label prediction in the original models. The performance with and without using Chaining is shown in Table \ref{Table:Chaining}. For the original MoE model, the number of mixtures is 16. For the Chaining MoE model, the number of mixture is 2, the number of stages is 8, and the predictions are projected into a 128 dimensions vector. For the original LSTM model, the number of mixtures is 8. For the Chaining LSTM model, the number of mixtures is 4, the number of stages is 2, and the dimension of projection is 200. For the original CNN model, the width of filter and the corresponding numbers of channel are $1 \times 512, 2 \times 512, 3 \times 1024$. For the Chaining CNN model, the corresponding parameters are $1 \times 128, 2 \times 128, 3 \times 256$, and the number of stages is 4. The parameters are chosen to make the original models have almost equal number of parameters with their Chaining counterparts.

\begin{table}[h] 
	\begin{center}
		\begin{tabular}{|l|l|c|c|}
			\hline
			Input Feature & Model & Original & Chaining \\
			\hline\hline
			Video-level, $\mu$ & MoE & 0.7965 & \textbf{0.8106} \\
			Frame Level, $\{x^v_{1:F_v}\}$  & LSTM & 0.8131 & \textbf{0.8172} \\
			Frame Level, $\{x^v_{1:F_v}\}$  & CNN & 0.7904 & \textbf{0.8179} \\
			\hline
		\end{tabular}
	\end{center}
	\caption{The performance (GAP) of Mixture-of-Experts, LSTM and CNN models with and without using Chaining structure.}
	\label{Table:Chaining}
\end{table}

\section{Temporal Multi-Scale Information} \label{Section:MultiScale}


Using information from different spatial scales have been discussed in many image analysis literatures. The main reason to do so is that the size of an object in an image would change with its distance to the observer. However, temporal information does not have the same rescaling effect. Therefore it may seems unnatural to model videos from different temporal scales.


However, we argue that temporal scales matters in video analysis. A task can be divided into several actions, and each action involves the interaction between certain objects. The label annotated to a video can be related to a concept at different temporal scale: an object, an action, or a task. 	 

\begin{figure}[t]
	\begin{center}
		\begin{subfigure}{0.45\textwidth}
			\includegraphics[width=\linewidth]{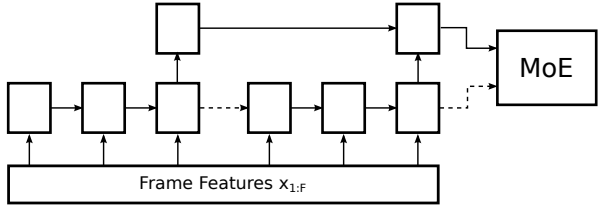}
			\caption{Temporal-pooling (with the dashes) and temporal-segment (without the dashes) LSTM model.} \label{Fig:MultiScale-pooling}
		\end{subfigure}
		\begin{subfigure}{0.45\textwidth}
			\vspace*{0.2in}
			\includegraphics[width=\linewidth]{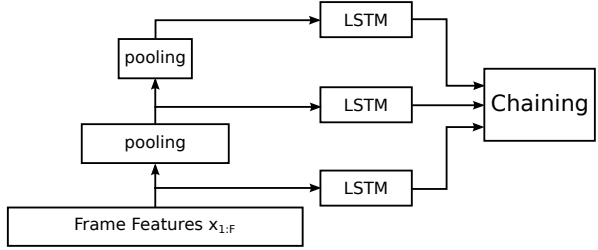}
			\caption{Multi-resolution LSTM model.} 
			\label{Fig:MultiScale-resolution}
		\end{subfigure}
		\begin{subfigure}{0.45\textwidth}
			\vspace*{0.2in}
			\includegraphics[width=\linewidth]{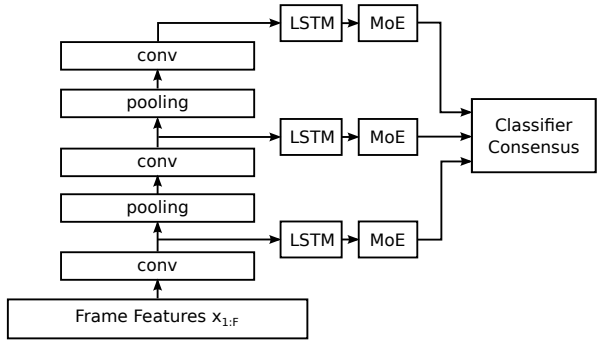}
			\caption{Multi-scale CNN-LSTM model.} 
			\label{Fig:MultiScale-cnnlstm}
		\end{subfigure}
	\end{center}
	\caption{Models that utilize temporal multi-scale information.}
	\label{Fig:MultiScale}
\end{figure}


One important observation is that video can be segmented into several clips. Each of these clips may contain content information of a certain aspect. We can make predictions based on each of these clips and then join them together to make more accurate predictions. The segmentation of video is done by either clustering the adjacent frames \cite{TemporalSeg2009Spriggs} or splitting the video into clips of equal durations \cite{TSN2016Wang}. We propose a temporal-segment LSTM model (Fig. \ref{Fig:MultiScale-pooling}), in which the video is split into equal-sized clips, each of which is modeled by an LSTM model. The models for different clips share the same parameters. The final state of each sequence are treated as another high-level sequence and modeled by another LSTM model. 

Temporal-pooling LSTM model (Fig. \ref{Fig:MultiScale-pooling}) is a multi-layer LSTM model in which we inserted temporal k-pooling layer between LSTM layers. It is similar to the temporal-segment LSTM model. The difference is whether to use the final state of the one clip as the initial state of the next clip in the first layer of the LSTM model.

Instead of directly segmenting videos into clips, we can aggregate the adjacent frames and gradually construct features containing long-range information. In multi-resolution LSTM model (Fig. \ref{Fig:MultiScale-resolution}), the original features are average pooled along the time dimension to get shorter sequences in which each frame covers longer time range. For each sequence generated in this way, a separate LSTM model is used to generate a sequence representation. The representations are then joined with a Chaining model. The representation from the highest-level goes into the first stage, while the one from the original sequence goes into the final stage of the Chaining model. This is due to the intuition that the representation of low-level sequences contain more refined features, while the high-level representation is close to an averaged view of the whole video.

We propose a novel temporal CNN-LSTM model (Fig. \ref{Fig:MultiScale-cnnlstm}) that utilize multi-scale information. This model shares the previous intuition that adjacent frames can be aggregated to generate features at different temporal scale. Instead of direct operating on the original feature, we use convolution layers to detect patterns in the low-level features and combine adjacent filter outputs by max-pooling along the time dimension. The filters used in a convolution layer are of the same length with the dimension of the feature, and their widths and channels can be varied. We use different LSTM models and Moe classifiers for the representation of the feature maps of different temporal scales. The predictions generated from features of different scales are combine using a consensus function. We find that averaging is a good consensus function, and maximum would lead to difficulties in convergence.


\begin{table}[h] 
	\begin{center}
		\begin{tabular}{|l|c|}
			\hline
			Model & GAP \\
			\hline\hline
			vanilla LSTM & 0.8131 \\
			\hline\hline
			temporal-pooling LSTM & 0.8085 \\
			temporal-segment LSTM & 0.8122 \\
			multi-resolution LSTM & 0.8148 \\
			multi-scale CNN-LSTM & 0.8204 \\
			\hline
		\end{tabular}
	\end{center}
	\caption{Performance of multi-scale models.}
	\label{Table:MultiScale}
\end{table}

Table \ref{Table:MultiScale} lists the performances of the models discussed in this section. The temporal-pooling LSTM model uses 2-pooling and has 4 LSTM layers (\#cells=1024) and 4 MoE mixtures. The temporal-segment LSTM model uses a duration of 10 frames for the clips and has 8 MoE mixtures and 2 LSTM layers, each of the layers has 1024 memory unit. In the multi-resolution LSTM model, we use 2-pooling to get shorter input sequences; the number of stages in Chaining is 4; the number of mixtures in MoE is 4; the projection dimension is 256; the LSTM models in it are 2-layer parallel LSTM (\#cells is 512 for video and 64 for audio). In the multi-scale CNN-LSTM model, the number of layers in CNN model is 4; the number of mixtures in MoE is 4; the LSTM models in it are 1-layer LSTM (\#cells=1024); in every layer of the CNN model, the width of filters and the corresponding number of channels are $1 \times 256, 2 \times 256, 3 \times 512$; the pooling layers in it are 2-pooling.

\section{Identifying Salient Frames with Attention} \label{Section:Attention}


\begin{figure*}[!htbp]
	\begin{center}
		\begin{subfigure}{0.4\textwidth}
			\includegraphics[width=\textwidth]{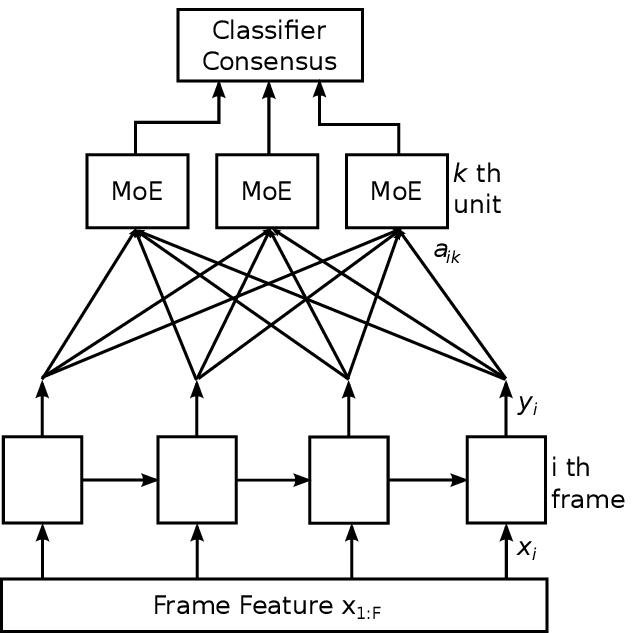}
			\caption{The multiple attention pooling scheme. The outputs of LSTM are pooled with multiple attention weights.} \label{Fig:Attention-pooling}
		\end{subfigure} 
		\hspace{0.05\textwidth}
		\begin{subfigure}{0.4\textwidth}
			\includegraphics[width=\textwidth]{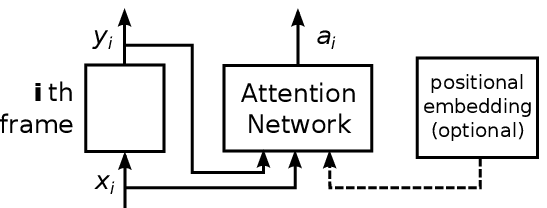}
			\caption{An attention network generates multiple attention weights for a frame based on the input feature and the LSTM output at that frame.} \label{Fig:Attention-network}
		\end{subfigure}
	\end{center}
	\caption{Multiple attention pooling model.}
	\label{Fig:Attention}
\end{figure*}

Not all the frames in a video are equally informative. There are many reasons that a frame might contribute little in video classification. Images that are too dark, too bright, or too blurry are hard to analysis for the image classification network. Therefore, such frames may not be able to provide useful semantic information. Also, sometimes a video contains title screens, credits or text. They might be useful if the frames are processed using optical character recognition (OCR). However, in this competition, the frames are preprocessed using an image classification network pre-trained with ImageNet dataset, which means the frame-level features may not contain the semantic information in these text. In addition, there are always frames irrelevant to the theme of the video. In talk show videos the content of frames will always be people talking while the topic of the talking is what really important. In documentaries there are often a lot of driving scenes which do not reflect the theme of the video.


We propose a multiple attention pooling scheme (multi-AP) for selecting salient frames in the video. As Fig. \ref{Fig:Attention} shows, an LSTM model is used to deal with the frame-level features. The outputs of the LSTM model are then aggregated by pooling over the time dimension. The aggregation is achieved by taking a weighted average over the outputs of the LSTM model, in which the weights are generated using an attention network, for which we use a fully connected layer. The aggregated feature is fed into an MoE model for label predictions. The attention pooling is repeated for multiple times and the predictions are combined using a classifier consensus function. For the choice of combining method, we find that maximum performs better than averaging as a consensus function. 

We use a fully connected layer with the softmax activation function as the attention network (Eq. \ref{Eq:AttentionNetwork}). It takes the input and output of the LSTM at a certain frame as the input. The attention network output $K$ groups of different attention weights.

\begin{equation}\label{Eq:AttentionNetwork}
	\begin{array}{c}
		\vspace{0.05in}
		e_{ik} = W_k [x_i; y_i]\\		
		\vspace{0.05in}
		a_{ik}  = \frac{exp(e_{ik})}{\sum_{i=1}^{F_v} exp(e_{ik})}\\
	\end{array}
\end{equation}

Each group of the weights are used to generate a group of aggregated features and prediction results. The results from all the groups are then combined by choosing the highest confidence predicted from all the models for each label $l$. The MoE models in Fig. \ref{Fig:Attention-pooling} share parameters with each other.

\begin{equation}\label{Eq:AttentionPooling}
	\begin{array}{c}
		\vspace{0.05in}
		z_k = \sum_{i=1}^{F_v} a_{ik} y_i\\
		\vspace{0.05in}
		p_k = MoE(z_k)\\
		\vspace{0.05in}
		p_l = \underset{k}{max} (p_{k,l})\\
	\end{array}
\end{equation}

\begin{figure*}[!htbp]
	\begin{center}
		\includegraphics[width=0.99\textwidth]{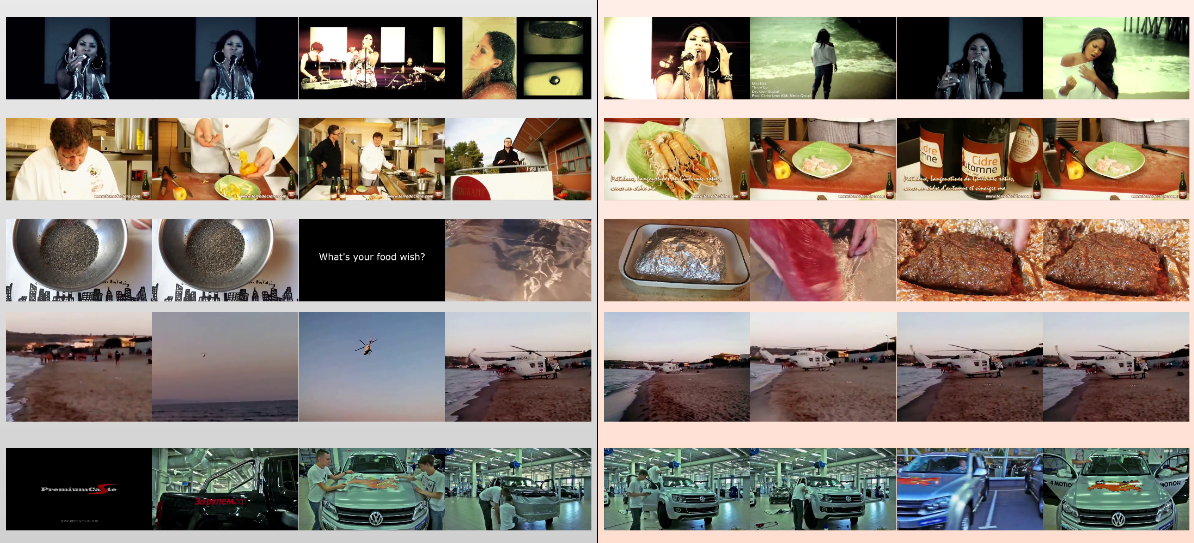}
	\end{center}
	\caption{Visualization of the attention weights in the multiple attention pooling model. The frames in each row comes from the same video. The left four in each row are the frames with the lowest weights, while the right four are the ones with the highest weights. }
	\label{Fig:AttentionPooling}
\end{figure*}

We present a few examples outside of the training set to show what kind of frames is highlighted by the attention network in the multi-AP model (Fig. \ref{Fig:AttentionPooling}). For each video, the frames with the lowest attention weights (left) and the ones with the highest attention weights (right) are presented. Since there are many groups of attention weight, the first group is used for this presentation. We have not observed any noticeable patterns in the inter-group differences of the weights.

In analysis of these examples, we find that there are some patterns about the attention weights. Title screens and the frames that is very dark tend to have low attention weights. Also, if the object of interest is too small or partly blocked, the weight of that frame tend to be low. 


Our work is different from a previous work on attention pooling \cite{AttentionPoolingCNN2016Er} in two ways. First, our model uses multiple groups of attention on the same output sequence and multiple classifiers for prediction. The multiple attention scheme may contribute to generating more stable classification result. Second, our model do an attention pooling over the outputs of LSTM model which is a global representation of the sequence while the previous work do a pooling over local CNN outputs. 

We also adopt a local attention pooling scheme that applies attention pooling over the input features (Eq. \ref{Eq:LocalAP}). The representation of the LSTM model is also used in the MoE model for prediction. 

\begin{equation}\label{Eq:LocalAP}
	\begin{array}{c}
		\vspace{0.05in}
		z = \sum_{i=1}^{F_v} a_{i} x_i\\
		\vspace{0.05in}
		p = MoE([y_{F_v}; z])\\
	\end{array}
\end{equation}


We compare the two attention pooling scheme with the baseline in Table \ref{Table:AttentionPooling}. The "local-AP LSTM" refers to the scheme using attention pooling over the input feature. The "multi-AP LSTM" refers to the multiple attention pooling scheme. The "positional multi-AP LSTM" model add an embedding for every frame position to the attention network on the basis of the multi-AP LSTM model.

The parameters for the models in Table \ref{Table:AttentionPooling} is as follows. The LSTM model in the four models share the same parameter (\#layers=2, \#cells=1024). The MoE models have 8 mixtures in the two multi-AP models, while the one in the local-AP model has 4 mixtures. The two multi-AP models both have 8 groups of attention weights. And the dimension of the positional embedding in the positional multi-AP model is 32.

\begin{table}[h] 
	\begin{center}
		\begin{tabular}{|l|c|}
			\hline
			Model & GAP \\
			\hline\hline
			vanilla LSTM & 0.8131 \\
			\hline\hline
			local-AP LSTM & 0.8133 \\
			multi-AP LSTM & 0.8157 \\
			positional multi-AP LSTM & 0.8169 \\
			\hline
		\end{tabular}
	\end{center}
	\caption{Performance (GAP) of attention models.}
	\label{Table:AttentionPooling}
\end{table}

\section{Bagging and Boosting} \label{Section:BaggingBoosting}

Bootstrap aggregating \cite{Bagging1996Breiman}, also called Bagging, is an ensemble method that creates many versions of a model and combines their results together. To create one version of a model, one applies sampling with replacement to get a subset of the original data, in which some original examples may not present or present more than once\cite{EnsembleMethods2012ZhouZH}. Training on different subset sampled from the original data would results in different models. The bagging algorithm is known for its ability to reduce the variance of a model. We apply Bagging to some of our models, and find that Bagging can generally boost the GAP performance by $0.6\% \sim 1.2\%$. The results are shown in Table \ref{Table:Bagging}.

\begin{table}[h] 
	\begin{center}
		\begin{tabular}{|l|l|c|c|}
			\hline
			Input Feature & Model & Original & Bagging \\
			\hline\hline
			Video-level  & Chaining & 0.8106 & \textbf{0.8225} \\
			\hline\hline
			Frame Level  & parallel LSTM & 0.8160 & \textbf{0.8216} \\
			Frame Level  & Chaining CNN & 0.8179 & \textbf{0.8258} \\
			Frame Level  & multi-AP LSTM & 0.8157 & \textbf{0.8244} \\
			\hline
		\end{tabular}
	\end{center}
	\caption{Performance (GAP) of bagging models.}
	\label{Table:Bagging}
\end{table}

Boosting \cite{Boosting1998breiman} is another way to create different versions of a model. Compared to Bagging which can be run in parallel, Boosting is a sequential ensemble method. The $(k+1)$th classifier is constructed considering the previous $k$ classifiers by re-sampling a distribution that highlights the misclassified examples. 

The re-sampling is often implemented by using weighted examples. If there are $N$ training examples and $L$ labels, the number of total classification results is $N \times L$. Most works on Boosting in multi-label classification use a weight $W^{N \times L}$ over $N$ samples and $L$ labels, which is computationally intractable when the vocabulary of label is large, as is the case in this competition. 

We adopt a per-example weighting scheme that assign a weight $W^N$ over the training samples. The weights are updated with Eq. \ref{Eq:Boosting}. In the equations, $W_{k,n}$ is the weight assigned to the $n$th example in the training of the $k$th classifier. ${Err}_{k,n}$ is the error rate of the $n$th example by evaluating the $k$th classifier. ${Err}_{k}$ is the average error rate of the $k$th classifier. And $Z_k$ is a coefficient that scales the average value of $W_{k,n}$ to 1. $\alpha$ is a parameter controlling the highlighting effect on the misclassified examples which we constantly set to 1 in our system.

\begin{equation}\label{Eq:Boosting}
	\begin{array}{l}
		\vspace{0.05in}
		W_{0,n} = 1.0\\
		\vspace{0.05in}
		W_{k+1,n} = \frac{N}{Z_k} W_{k,n} \exp(\alpha r_{k} {Err}_{k,n}) \\
		\vspace{0.05in}
		in ~ which, \\
		\vspace{0.05in}
		r_k = \log(\frac{1.0 - {Err}_k}{{Err}_k}) \\
		\vspace{0.05in}
		{Err}_k = \frac{1}{N} \sum_{n=1}^{N} {Err}_{k,n}\\
		\vspace{0.05in}
		{Err}_{k,n} \in [0,1]\\
		\vspace{0.05in}
		Z_k = \sum_{n=1}^{N} W_{k,n} \exp(\alpha r_{k} {Err}_{k,n})
	\end{array}
\end{equation}

In this algorithm, the weights of the misclassified examples are increased in the following classifiers. However, in multi-label classification, misclassification is hard to define. We choose the Precision Equal Recall Rate (PERR) as the implementation of error rate, since it is both per-example evaluated and in coordinate with the GAP score in most models (from empirical observation). We also clip the weights to 5 to ensure that the algorithm would not place all the weights on a few formerly misclassified examples, since they may not be visually classifiable. 

\begin{table}[h] 
	\begin{center}
		\begin{tabular}{|l|c|c|}
			\hline
			Model & Original & Boosting \\
			\hline\hline
			Chaining (Video) & 0.8106 & \textbf{0.8218} \\
			\hline\hline
			parallel LSTM & 0.8160 & \textbf{0.8218} \\
			Chaining CNN & 0.8179 & \textbf{0.8242} \\
			multi-AP LSTM & 0.8157 & \textbf{0.8246} \\
			\hline
		\end{tabular}
	\end{center}
	\caption{Performance (GAP) of boosting models.}
	\label{Table:Boosting}
\end{table}

The comparison between the single models and their Boosting counterparts are shown in Table \ref{Table:Boosting}. The Boosting algorithm generates a performance boost similar as Bagging does. 

\section{Cascade Classifier Training} \label{Section:Cascade}

Adding models into the ensemble would usually make the performance better. However, with the number of models in the ensemble model increasing, the gain from newly added models tends to diminish, especially if one add models that are similar to the existing models. We address this problem by using cascade training in which the predictions of other models as a part of its input of the model during training. 

\begin{figure}[!htbp]
	\begin{center}
		\includegraphics[width=0.4\textwidth]{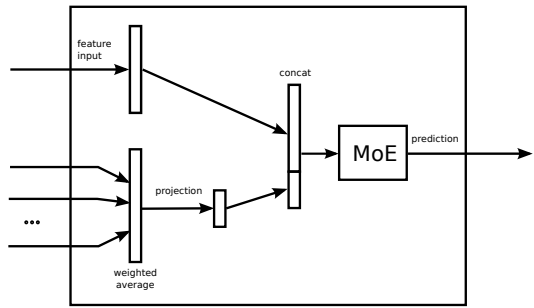}
	\end{center}
	\caption{Cascade layer as a replacement for MoE in cascade classifier training.}
	\label{Fig:Cascade}
\end{figure}

The structure of cascade layer is shown in Fig. \ref{Fig:Cascade}. The predictions from the other models are averaged to generate an averaged prediction, which is then projected to a feature of low dimension. All the models that have MoE as their last layer can be modified into a cascade model by having their MoE layer replaced by the cascade layer.

\section{Stacking Methods} \label{Section:Stacking}

Stacking is an ensemble method that uses a machine learning model to combine the predictions of many models into one prediction. We use the "train2" set (Section \ref{Section:Introduction}) as the training set for stacking. The trivial case of stacking is to simply average all the predictions. If there are $M$ models and $L$ labels in total, simple averaging the predictions of all models can be written as Eq. \ref{Eq:Stacking-simple}.

\begin{equation}\label{Eq:Stacking-simple}
	p_l = \frac{1}{M} \sum_{m=1}^{M} p_{m,l}
\end{equation}

Linear weighted averaging (Eq. \ref{Eq:Stacking-linear}) is also a simple scheme. This method has $M$ weights, one for each model in the ensemble.

\begin{equation}\label{Eq:Stacking-linear}
	p_l = \frac{1}{M} \sum_{m=1}^{M} w_m p_{m,l}, \hspace{0.1in} s.t. \sum_{m=1}^{M} w_m = 1, w_m > 0 
\end{equation}

For multi-label classification, there are $L$ predictions for every example. Some models might perform better than the others on a subset of the labels. Therefore, we could extend linear weighted averaging to every class. We denote this method class-wise weighted averaging (Eq. \ref{Eq:Stacking-class}). 

\begin{equation}\label{Eq:Stacking-class}
	p_l = \frac{1}{M} \sum_{m=1}^{M} w_{m,l} p_{m,l}, \hspace{0.1in} s.t. \sum_{m=1}^{M} w_{m,l} = 1, w_{m,l} > 0 
\end{equation}

While the ensemble methods often deal with the predictions of individual models, the original input may also help to decide which models to trust. The winning team of Netflix Prize proposes the Feature-Weighted Linear Stacking \cite{FWLS2009Sill} to utilize the information in meta-features. Although there are no useful meta-feature in the Youtube-8M dataset, we consider the averaged frame input and the averaged predictions from individual models as useful indicators. We propose the attention weighted stacking to utilized these information.

The attention weights $\alpha$ (Eq. \ref{Eq:Stacking-attentionweights}) are generated with a fully connected layer with softmax as the activation function and averaged input $\bar{x}$ and prediction $\bar{p}$ as the input of the layer. 

\begin{equation}\label{Eq:Stacking-attentionweights}
	\begin{array}{ccl}
		\vspace{0.1in}
		\alpha & = & \sigma(V [\bar{x};\bar{p}])\\
		\vspace{0.1in}
		\bar{x} & = & \frac{1}{F_v} \sum_{i=1}^{F_v} x_i\\
		\vspace{0.1in}
		\bar{p} & = & \frac{1}{M} \sum_{m=1}^{M} p_m\\
	\end{array}
\end{equation}

\begin{equation}\label{Eq:Stacking-attentionstacking}
	\begin{array}{ccl}
		\vspace{0.1in}
		p_l & = & \frac{1}{M} \sum_{m=1}^{M} w_{m,l} p_{m,l}\\
		\vspace{0.1in}
		w_{m,l} & = & \frac{\exp(e_{m,l})}{\sum_{m=1}^{M} \exp(e_{m,l})}\\
		\vspace{0.1in}
		e & = & \sum_{k=1}^{K} \alpha_k (A_k^T B_k + a_k b_k^T + c_k)\\
		& & A_k \in R^{D \times M}, ~ B_k \in R^{D \times L}\\
		s.t. & & a_k \in R^{M \times 1}, ~ b_k \in R^{L \times 1}\\
		& & c_k \in R
	\end{array}
\end{equation}

The predictions are then weighted averaged as shown in Eq. \ref{Eq:Stacking-attentionstacking}. The weight is a mixture of $K$ low-rank matrices (rank $D$) in which the matrices are selected by the output of the attention network (Eq. \ref{Eq:Stacking-attentionweights}). In the attention weighted stacking, $V, A, B, a, b, c$ are trainable parameters.

We perform a comparison among the stacking methods on an ensemble of 74 models (model \#1 - \#74 in Appendix \ref{Appendix}). The number of attention weights is 16 and the rank of the matrix components is 4 in the attention weighted stacking method. The comparison shows that the attention weighted stacking performs better than the other methods (Table \ref{Table:Stacking}).

\begin{table}[!htbp]
	\begin{center}
		\begin{tabular}{|l|c|}
			\hline
			Stacking method  & GAP \\
			\hline\hline
			Simple Averaging  & 0.8436 \\
			Linear Weighted Averaging & 0.8449 \\
			Class-wise Weighted Averaging  & 0.8453 \\
			Attention Weighted Stacking  & \textbf{0.8458} \\
			\hline
		\end{tabular}
	\end{center}
	\caption{Performance (GAP) of different stacking models on the entire set of models (74 models). The GAP scores in this table is evaluated with the blind test set (the private leaderboard score in the competition).}
	\label{Table:Stacking}
\end{table}

\section{Distilling Knowledge from Ensemble} \label{Section:Distill}

Ensemble models might perform much better than even the best single model. However, in real world systems, the memory and computation resources are often too limited for ensemble models to deploy. Distillation \cite{Distill2015hinton} is a training scheme that aims to compress the knowledge in ensemble model to a single model. Another motivation of using distillation is the observation that the labels are often wrong or incomplete, especially when the vocabulary of label is large. 

We adopt a training scheme that uses the predictions of ensemble $\hat{p}$ as a soft target along with the real target $l$ during training. The new loss function is a weighted average of two losses, one for each target (Eq. \ref{Eq:Distillation}).

\begin{equation}\label{Eq:Distillation}
	\begin{array}{c}
		\vspace{0.1in}
		\hat{L} = (1-\lambda) ce(p, l) + \lambda ce(p, \hat{p})
	\end{array}
\end{equation}

\begin{table}[h] 
	\begin{center}
		\begin{tabular}{|l|c|c|}
			\hline
			Model & Original & Distillation \\
			\hline\hline
			Chaining (Video) & 0.8106 & \textbf{0.8169} \\
			parallel LSTM & 0.8160 & \textbf{0.8237} \\
			Chaining CNN & 0.8179 & \textbf{0.8266} \\
			Chaining LSTM & 0.8172 & \textbf{0.8291} \\
			Multi-scale CNN-LSTM & 0.8204 & \textbf{0.8258} \\
			\hline
		\end{tabular}
	\end{center}
	\caption{Performance (GAP) of single models trained using distillation.}
	\label{Table:Distillation}
\end{table}

We use the predictions of model \#75 (Appendix \ref{Appendix:Ensemble}) as the soft target in training and train several models with the same parameters as the original models. The comparison is listed in Table \ref{Table:Distillation}. The comparison shows that distillation can greatly boost single model performance (often better than Bagging).

\section{Summary} \label{Section:Conclusion}

In the previous sections, we review our system, present our intuition, and show how we address the problems in multi-label video classification. We find that attention pooling and temporal multi-scale information is very important for video sequence modeling. We also propose a network structure for large-vocabulary multi-label classification. We review our work in ensemble methods such as Bagging, Boosting, Cascade, Distillation and Stacking. We propose a stacking network that uses attention to weight models. Our system road-map is shown in Table \ref{Table:Summary}.

\begin{table}[h] 
	\begin{center}
		\begin{tabular}{|l|c|}
			\hline
			Changes & GAP \\
			\hline\hline
			27 single models  & 0.8425 \\
			+ 11 bagging \& boosting models & 0.8435 \\
			+ 8 distillation models  & 0.8437 \\
			+ 11 cascade models & 0.8451 \\
			+ 17 more cascade models  & 0.8453 \\
			Attention Weighted Stacking  & \textbf{0.8458}\\
			\hline
		\end{tabular}
	\end{center}
	\caption{Performance (GAP) of ensemble models. The stacking method used for the first 5 results is class-wise weighted model. The ensemble in each row include all the changes in the rows above it. The results reported in this table are evaluated on the blind test set (the private leaderboard score in the competition).}
	\label{Table:Summary}
\end{table}

\newpage
{\small
\bibliographystyle{ieee}
\bibliography{egbib}
}

\newpage
\begin{appendices}
	\section{Final Submission} \label{Appendix}
	
		In this section we list all the components included in our final submission. We apply a stacking method called the Attention Weighted Average (Section \ref{Section:Stacking}) to merge the result of these models instead of carrying on careful model selection. Therefore it might be possible that a subset of these models could reach equal or better performance. The parameters are for a quick grasp of the structure of the models. Refer to our code \footnote{https://github.com/wangheda/youtube-8m} if you are interested in the implementation details. The GAP scores reported in the appendix are evaluated with the "validate2" set if not otherwise specified.
		
		\subsection{Video-level Models} \label{Appendix:Video}
		
		1. Chaining model (Section \ref{Section:Chaining}). During training, if the loss of the current batch is less than 10, use the most confident tag predicted as a soft target for every example in the batch. \#mixture of MoE is 4. \#stage of Chaining is 3. projection dimension in Chaining is 100. GAP = 0.8067.
		
		2. Chaining model. \#mixture of MoE is 2. \#stage of Chaining is 8. projection dimension in Chaining is 128. GAP = 0.8106.
		
		\subsection{Baseline Models} \label{Appendix:Baseline}
		
		If not otherwise specified, the memory cells of the final state are used for the input of the MoE model in the baseline models. 
		
		3. LSTM model (Section \ref{Section:Baseline}). \#cell of LSTM is 1024. \#layer of LSTM is 2. \#mixture of MoE is 8. GAP = 0.8131.
		
		4. LSTM model. \#cell of LSTM is 2048. \#layer of LSTM is 2. \#mixture of MoE is 4. GAP = 0.8152.
		
		5. LSTM model of 2 layers. The first layer is bi-directional. The second layer is uni-directional. \#cell of LSTM is 1024,. \#mixture of MoE is 4. GAP = 0.8105.
		
		6. LSTM-S model (Section \ref{Section:Baseline}). \#cell of LSTM is 1024. \#layer of LSTM is 1. \#mixture of MoE is 8. GAP = 0.8123.
		
		7. LSTM-A model (Section \ref{Section:Baseline}).  \#cell of LSTM is 1024. \#layer of LSTM is 1. \#mixture of MoE is 8. GAP = 0.8131.
		
		8. LSTM-A model of 2 layers. The first layer is a forward LSTM-A model. The second layer is a backward LSTM-A model that takes the original input and the output of the first layer as input. \#cell of LSTM is 1024. \#layer of LSTM is 1. \#mixture of MoE is 8. GAP = 0.8131.
		
		9. Parallel LSTM model (Section \ref{Section:Baseline}). \#cell of LSTM is 1024. \#layer of LSTM is 2. \#mixture of MoE is 8. GAP = 0.8161.
		
		10. Parallel LSTM model. The gated outputs of the final state are used for the input of the MoE model. \#cell of LSTM is 1024. \#layer of LSTM is 2. \#mixture of MoE is 8. GAP = 0.8160.
		
		11. LSTM model with data augmentation by random sampling 50\% of the frames.  \#cell of LSTM is 1024. \#layer of LSTM is 2. \#mixture of MoE is 8. GAP = 0.8137.
		
		12. CNN-LSTM model. The CNN model is described in section \ref{Section:Chaining}. \#layer of CNN is 1. The widths of filter are 1, 2 and 3. The corresponding \#channels of filters are 1024, 1024, and 1024. \#cell of LSTM is 1024. \#layer of LSTM is 2. \#mixture of MoE is 8. GAP = 0.8103.
		
		\subsection{Temporal Multi-Scale Models} \label{Appendix:Multiscale}
		
		13. Temporal-segment LSTM model (Section \ref{Section:MultiScale}). The duration of each clip is 10 frames. \#cell of LSTM is 1024. \#layer of LSTM is 2. \#mixture of MoE is 8. GAP = 0.8122.
		
		14. Temporal-pooling LSTM model (Section \ref{Section:MultiScale}). The model has 2-pooling between LSTM layers. \#cell of LSTM is 1024. \#layer of LSTM is 4, \#mixture of MoE is 8. GAP = 0.8085.
		
		15. Multi-resolution LSTM model (Section \ref{Section:MultiScale}). The model has parallel LSTM for sequence modeling. The model has 2-pooling to get shorter representation of input. \#cell of LSTM is 512 for video and 64 for audio. \#layer of LSTM is 2. \#stage of Chaining is 4. The dimension of projection in Chaining is 256. \#mixture of MoE is 4. GAP = 0.8149.
		
		16. Multi-scale CNN-LSTM model (Section \ref{Section:MultiScale}). The widths of filter are 1, 2 and 3. The corresponding \#channels of filters are 256, 256, and 512. \#layer in CNN is 4.  \#layer of LSTM is 1. \#cell of LSTM is 1024. \#mixture of MoE is 4. GAP = 0.8204.
		
		17. Multi-scale CNN-LSTM model. All the parameters are the same as model \#16, except that the type of the LSTM model is LSTM-S (as in model \#6). GAP = 0.8147.
		
		\subsection{Chaining Models} \label{Appendix:Chaining}
		
		18. Chaining CNN model (Section \ref{Section:Chaining}). The widths of filter are 1, 2 and 3. The corresponding \#channels of filters are 128, 128, and 256. The dimension of projection in Chaining is 256. \#mixture of MoE is 4. \#stage of Chaining is 4.  GAP = 0.8179.
		
		19. Chaining Deep CNN model. In this model a 3-layer CNN model is used. The averaged input feature and the 3 max-pooled feature maps from the 3 layers of CNN are combined using a 4-stage Chaining model.  The widths of filter are 1, 2 and 3. The corresponding \#channels of filters are 128, 128, and 256. The dimension of projection in Chaining is 256. \#mixture of MoE is 4. GAP = 0.8155.
		
		20. Chaining LSTM-CNN model. This model uses a cascade of parallel LSTM and CNN model as the sub-model in Chaining.\#layer of LSTM is 1. \#cell of LSTM is 1024 for video and 128 for audio. The widths of filter are 1, 2 and 3. The corresponding \#channels of filters are 128, 128, and 256. The dimension of projection in Chaining is 128. \#stage in Chaining is 3. \#mixture of MoE is 4. GAP = 0.8122.
		
		21. Chaining LSTM model (Section \ref{Section:Chaining}). \#layer of LSTM is 2. \#cell of LSTM is 1024. The dimension of projection in Chaining is 200. \#stage in Chaining is 2. \#mixture of MoE is 4. GAP = 0.8172.
		
		22. Chaining LSTM model. The different stages in Chaining model uses a shared input which is the cell memory of the final state of an LSTM model. \#layer of LSTM is 2. \#cell of LSTM is 1024. The dimension of projection in Chaining is 256. \#stage in Chaining is 2. \#mixture of MoE is 4. GAP = 0.8162.
		
		\subsection{Attention Pooling Models} \label{Appendix:Attention}
		23. Local attention pooling LSTM model (Section \ref{Section:Attention}). \#layer of LSTM is 2. \#cell of LSTM is 1024. GAP = 0.8133.
		
		24. Attention pooling LSTM model that has one LSTM model to generate the attention weights for pooling over the output of another LSTM model. \#layer of LSTM is 1. \#cell of LSTM is 1024. \#mixture of MoE is 8. GAP = 0.8088.
		
		25. Multiple attention pooling LSTM model (Section \ref{Section:Attention}). \#layer of LSTM is 2. \#cell of LSTM is 1024. \#mixture of MoE is 4. \#group of attention weights is 8. GAP = 0.8157.
		
		26. Multiple attention pooling LSTM model that has one LSTM model to generate the attention weights for pooling over the output of another LSTM model. \#layer of LSTM is 2. \#cell of LSTM is 1024. \#mixture of MoE is 4. \#group of attention weights is 8. GAP = 0.8081.
		
		27. Multiple attention pooling LSTM model with positional embedding (Section \ref{Section:Attention}). \#layer of LSTM is 2. \#cell of LSTM is 1024. \#mixture of MoE is 4. The dimension of positional embedding is 32. \#group of attention weights is 8. GAP = 0.8169.
		
		\subsection{Bagging and Boosting Models} \label{Appendix:BaggingBoosting}
		
		28. Bagging model of 8 versions of model \#2, combined with simple averaging. GAP = 0.8225.
		
		29. Ensemble of 4 Chaining models. The first is the normal Chaining model with video-level input. The second is a Chaining model with weighted cross entropy loss. The third is a Chaining model that predict label and top-level verticals at the same time. The fourth is a Chaining model that predict the infrequent labels with softmax function. All 4 models share the same parameters.  \#mixture of MoE is 4.  \#stage in Chaining is 3. The dimension of projection in Chaining is 100. GAP = 0.8216.
		
		30. Boosting model of 8 versions of model \#2, combined with class-wise weighted averaging. Remove the examples with too high weights from training set, since it may be hopeless to predict these examples correctly. GAP = 0.8213.
		
		31. Boosting model of 8 versions of model \#2, combined with class-wise weighted averaging, without weight clipping. GAP = 0.8198.
		
		32. Boosting model of 8 versions of model \#2, combined with class-wise weighted averaging, with weight clipping. GAP = 0.8218.
		
		33. Bagging model of 8 versions of model \#18, combined with class-wise weighted averaging. GAP = 0.8258.
		
		34. Boosting model of 8 versions of model \#18, combined with class-wise weighted averaging, with weight clipping. GAP = 0.8246.
		
		35. Bagging model of 8 versions of model \#25, combined with class-wise weighted averaging. GAP = 0.8244.
		
		36. Boosting model of 8 versions of model \#25, combined with attention weighted stacking, with weight clipping. GAP = 0.8242.
		
		37. Bagging model of 8 versions of model \#10, combined with simple averaging. GAP = 0.8216.
		
		38. Boosting model of 8 versions of model \#10, combined with class-wise weighted averaging, with weight clipping. GAP = 0.8218.
		
		\subsection{Distillation Models} \label{Appendix:Distillation}
		
		The models in this section is trained using distillation (Section \ref{Section:Distill}). During training, the predictions from model \#75 are used as soft targets.
		
		39. Model \#2 with distillation training. GAP = 0.8169.
		
		40. Model \#18 with distillation training. GAP = 0.8266.
		
		41. Model \#20 with distillation training. GAP = 0.8259.
		
		42. Model \#10 with distillation training. GAP = 0.8237.
		
		43. Model \#21 with distillation training. GAP = 0.8291.
		
		44. Model \#16 with distillation training. GAP = 0.8258.
		
		45. Bagging model of 4 versions of model \#2 with distillation training. GAP = 0.8249.
		
		46. Boosting model of 4 versions of model \#2 with distillation training. GAP = 0.8254.

		\subsection{Cascade Models} \label{Appendix:Cascade} 
		
		The models in this section are models using the prediction of other models as part of the input (Section \ref{Section:Cascade}). 
		
		47. Model \#2 with cascade training using the prediction of model \#75 as part of the input. GAP = 0.8231.
		
		48. Model \#6 with cascade training using the prediction of model \#75 as part of the input. GAP = 0.8245.
		
		49. Model \#18 with cascade training using the prediction of model \#75 as part of the input. GAP = 0.8268.
		
		50. Model \#25 with cascade training using the prediction of model \#75 as part of the input. GAP = 0.8267.
		
		51. Chaining CNN model with cascade training using the prediction of model \#75 as part of the input. The widths of filter are 1, 2 and 3. The corresponding \#channels of filters are 256, 256, and 512. The dimension of projection in Chaining is 256. \#mixture of MoE is 4. \#stage of Chaining is 2. GAP = 0.8214.
		
		52. Model \#3 with cascade training using the prediction of model \#75 as part of the input. GAP = 0.8267.
		
		53. Model \#16 with cascade training using the prediction of model \#75 as part of the input. \#stage of Chaining is 2. GAP = 0.8214.
		
		54. Model \#16 with cascade training using the prediction of model \#75 as part of the input. \#stage of Chaining is 4. GAP = 0.8266.
		
		55. Model \#20 with cascade training using the prediction of model \#75 as part of the input.  GAP = 0.8265.
		
		56. Model \#6 with cascade training using the prediction of model \#75 as part of the input. The prediction of model \#6 for infrequent labels and the prediction of the cascade model for frequent ones are joined together as the final prediction. GAP = 0.8228.
		
		57. Model \#2 with cascade training using the prediction of model \#76 as part of the input. GAP = 0.8202.
		
		58. Model \#25 with cascade training using the prediction of model \#76 as part of the input. GAP = 0.8254.
		
		59. Model \#20 with cascade training using the prediction of model \#76 as part of the input.  GAP = 0.8250.
		
		60. Model \#6 with cascade training using the prediction of model \#76 as part of the input. GAP = 0.8245.
		
		61. Model \#7 with cascade training using the prediction of model \#76 as part of the input. GAP = 0.8247.
		
		62. Model \#21 with cascade training using the prediction of model \#76 as part of the input. GAP = 0.8251.
		
		63. Model \#16 with cascade training using the prediction of model \#76 as part of the input. GAP = 0.8258.
		
		64. Model \#10 with cascade training using the prediction of model \#76 as part of the input. GAP = 0.8248.
		
		65. Model \#10 with cascade training using the prediction of model \#76 as part of the input. The prediction of model \#76 is also used to up-sample the misclassified samples as in the Boosting model. GAP = 0.8218.
		
		66. Model \#2 with cascade training using the prediction of model \#76 as part of the input. GAP = 0.8181.
		
		67. Model \#2 with cascade training using the predictions of model \#76 and \#66 as part of the input. GAP = 0.7989.
		
		68. Model \#2 with cascade training using the predictions of model \#76, \#66, and \#67 as part of the input. GAP = 0.7849.
		
		69. Model \#2 with cascade training using the predictions of model \#76, \#66, \#67, and \#68 as part of the input. GAP = 0.7753.
		
		70. Model \#2 with cascade training using the predictions of model \#76, \#66, \#67, \#68, and \#69 as part of the input. GAP = 0.7833.
		
		71. Model \#2 with cascade training using the predictions of model \#76 as part of the input. GAP = 0.8183.
		
		72. Model \#2 with cascade training using the predictions of model \#76 and \#71 as part of the input. GAP = 0.8085.
		
		73. Model \#2 with cascade training using the predictions of model \#76, \#71, and \#72 as part of the input. GAP = 0.8111.
		
		74. Model \#2 with cascade training using the L1-normalized prediction of model \#76 as part of the input. GAP = 0.8227.
		
		\subsection{Ensemble Models} \label{Appendix:Ensemble}
		
		In this section we lists the performance of some of the ensemble models. Alongside the GAP score on "validate2" set, we also report the GAP score on the blind test set (Private Leader\-Board in the competition).
		
		75. Ensemble of 4 single models. Model \#2, \#10, \#18, and \#25 with a class-wise weighted stacking gets a GAP of 0.8373. Private LB GAP = 0.83703.
		
		76. Ensemble of 8 single models. Model \#2, \#10, \#18, \#25, \#47, \#48, \#49, and \#50 with a class-wise  weighted stacking gets a GAP of 0.8397. Private LB GAP = 0.83941.
		
		77. Ensemble of 27 single models. Model \#1 - \#27 with a class-wise weighted stacking gets a GAP of 0.8427. Private LB GAP = 0.84250.
		
		78. Ensemble of 57 models. Model \#1 - \#57 with an attention weighted stacking (\#rank=3, \#attention=16) gets a GAP of 0.8459. Private LB GAP = 0.84561.
		
		79. Ensemble of 74 models. Model \#1 - \#74 with an attention weighted stacking (\#rank=4, \#attention=16) gets a GAP of 0.8462. Private LB GAP = 0.84583. 
		
		80. Ensemble of model \#78 and \#79 by getting the most confident 20 tags from each model and averaging the confidence scores. This is done directly on the test set without training. Private LB GAP = 0.84590. This is our final submission. 
		
	\section{Source Code} \label{Appendix:SourceCode}
	
		The source code of our solution is made public on GitHub. Visit https://github.com/wangheda/youtube-8m for the implementation details of our models.

\end{appendices}

\end{document}